\documentclass[10pt,twocolumn,letterpaper]{article}

\usepackage{wacv}      

\usepackage{graphicx}
\usepackage{amsmath}
\usepackage{bm}
\usepackage{bbm}
\usepackage{amssymb}
\usepackage{booktabs}
\usepackage[ruled,linesnumbered]{algorithm2e}

\usepackage[pagebackref,breaklinks,colorlinks]{hyperref}

\usepackage[capitalize]{cleveref}
\crefname{section}{Sec.}{Secs.}
\Crefname{section}{Section}{Sections}
\Crefname{table}{Table}{Tables}
\crefname{table}{Tab.}{Tabs.}

\def\wacvPaperID{2054} 
\def\confName{WACV}
\def\confYear{2025}

\begin{document}

\title{GS-Matching: Reconsidering Feature Matching task in Point Cloud Registration}

\author{Yaojie Zhang, Tianlun Huang, Weijun Wang\\
Shenzhen Institute of Advanced Technology, Chinese Academy of Sciences\\
Shenzhen, 518055, China\\
{\tt\small yj.zhang1@siat.ac.cn}
\and
Wei Feng\\
Guangdong Provincial Key Laboratory of Construction Robotics and Intelligent Construction\\
Shenzhen, 518055, China\\
{\tt\small wei.feng@siat.ac.cn}
}
\maketitle

\begin{abstract}
   Traditional point cloud registration (PCR) methods for feature matching often employ the nearest neighbor policy. This leads to many-to-one matches and numerous potential inliers without any corresponding point. Recently, some approaches have framed the feature matching task as an assignment problem to achieve optimal one-to-one matches. We argue that the transition to the Assignment problem is not reliable for general correspondence-based PCR. In this paper, we propose a heuristics stable matching policy called GS-matching, inspired by the Gale-Shapley algorithm. Compared to the other matching policies, our method can perform efficiently and find more non-repetitive inliers under low overlapping conditions. Furthermore, we employ the probability theory to analyze the feature matching task, providing new insights into this research problem. Extensive experiments validate the effectiveness of our matching policy, achieving better registration recall on multiple datasets. 
\end{abstract}

\section{Introduction}
\label{sec:intro}

Point cloud registration (PCR) stands as a vital and foundational challenge within 3D computer vision, with broad implications for Simultaneous Localization and Mapping (SLAM) \cite{xu2022fast}, 3D object detection \cite{guo20143d}, and robotics applications \cite{kostavelis2015semantic}. When dealing with two 3D scans capturing the same object or scene, PCR aims to determine a precise six-degree-of-freedom (6-DoF) pose transformation to align input point clouds accurately. The conventional correspondence-based PCR involves two key steps: feature correspondence establishment and rigid transformation estimation. Specifically, feature correspondence establishment initially with extracting a feature descriptor for each keypoint, followed by establishing correspondences in feature space.

Given feature descriptors, putative feature correspondences by default are generated by finding the nearest neighbor in the feature space. Under this matching policy, each source keypoint is matched with the target keypoint possessing the highest feature similarity score. However, this policy neglects the one-to-one matching principle and leads to the many-to-one matching problem \cite{zhang2022searching}. 

Although 3D point cloud registration task has received widespread attention in past decades, the policy to generate putative feature correspondences from feature descriptors has not been investigated widely.  Previous works, such as \cite{choy2019fully,zeng20173dmatch, yang2020teaser,bai2021pointdsc,chen2022sc2,zhang20233d} only focus on the impact of different descriptor types and outlier rejection algorithms. The matching policy to generate correspondences is not considered. Some deep-learned approaches \cite{sarlin2020superglue,10076895,yu2021cofinet, zhang2022searching} treat the feature matching as an Assignment Problem to achieve optimal one-to-one matches. However, this transition is not reliable for partial overlap PCR. We will also explain the reason and show the experiment result in the following \cref{subsec:AP_problem} and \cref{sec:Analysis_experiments}. In this paper, our primary focus is on examining the impact of the matching policy on registration performance. We consider a good generate policy should ensure real-time performance itself, generate as many inliers as possible, and be compatible with current mainstream correspondence-based PCR frameworks. Guided by these principles, we propose a heuristically stable matching policy called GS-Matching which is inspired by the Gale-Shapley algorithm. Additionally, we analyze the feature matching task using the probability theory and provide a novel perspective.  
In summary, our main contributions are as follows:
\begin{enumerate}
    \item A heuristically stable matching policy called GS-Matching is proposed. The proposed GS-Matching can find more reliable inliers compared to other matching policies thereby improving the registration performance.
    \item Proposal of a method, informed by probability theory, to reduce point cloud size without compromising registration performance.
    \item Validation of GS-Matching's suitability for general correspondence-based PCR through extensive experiments, showcasing its performance enhancement across diverse datasets.
\end{enumerate}

\section{Related Work}
Correspondence-based PCR involves two major research topics: 3D feature matching and outlier rejection.
\subsection{3D Feature Matching}
3D feature matching aims to establish initial correspondences based on feature descriptors. Feature descriptors typically take the form of vectors or histograms, describing the surrounding information of keypoints. Hand-crafted descriptors have been comprehensively reviewed in \cite{guo2016comprehensive}. Spatial distribution histogram based descriptors \cite{johnson1999using, frome2004recognizing, tombari2010unique, guo2015novel} represented the local surface according to the spatial distributions (e.g., coordinates) information. Geometric attribute histogram based descriptors \cite{chen20073d, rusu2008aligning, rusu2009fast, salti2014shot} utilized the geometric attributes (e.g., normals, curvatures) of points.
Recently, deep learning techniques have been introduced to learn 3D local descriptors. The pioneering 3DMatch \cite{zeng20173dmatch} employed a Siamese Network for extracting local descriptors. Some recent networks \cite{ao2021spinnet,yang2018foldingnet, choy2019fully, yew20183dfeat, bai2020d3feat, huang2021predator} aimed to extract more descriptive and robust descriptors by designing new network architecture or combining geometric information.

These studies emphasized the extraction of highly distinctive features, where a single feature can be accurately matched with a high probability against numerous unrelated features. However, the policy to generate correspondences lacks study and by default is finding the nearest neighbor in feature space. Despite significant advancements in deep learning-based 3D feature matching technology, these methods are still prone to generate correspondences with extremely high outlier rates, posing a major challenge for correspondence-based PCR. As mentioned in \cref{sec:intro}, the nearest neighbor policy results in massive repeat matching points which could be one of the reasons for the low inlier ratio issue. 

\subsection{Outlier rejection}
The outlier rejection method aims to remove outliers from initial correspondences to accurately estimate the geometric model for input point clouds. The widely used RANSAC \cite{fischler1981random} employed a generation-and-verification pipeline for robust outlier removal. However, a common issue of the RANSAC and its variants \cite{barath2018graph,quan2020compatibility,barath2019magsac, chum2005matching} is the challenge posed by low inlier ratios, making it difficult to find a reasonable model efficiently. To improve the RANSAC performance, GORE \cite{bustos2017guaranteed} reduced the size of correspondences by rejecting most true outliers. Due to the time complexity of the RANSAC method, various RANSAC-free methods have been explored. 
One of the representative works is \cite{leordeanu2005spectral}, which proposed the spatial compatibility technique. The spatial compatibility measure-based method utilizes correspondence-wise constraints to generate an adjacency matrix and transfer the outlier rejection problem to a maximum clique problem in graph theory. 
Clipper \cite{lusk2021clipper} and Teaser \cite{yang2020teaser} introduced a graph-theoretic framework to achieve robust data association. SC2 \cite{chen2022sc2} presented a second-order spatial compatibility measure allowing for more distinctive clustering compared to the original measure. MAC \cite{zhang20233d} relaxed the maximum clique constraint to a maximal clique constraint which can mine more local information in a graph. Global optimal approaches can also achieve outlier robustness registration. Go-ICP \cite{yang2015go} used a branch-and-bound (BnB) scheme for globally optimal registration. FGR \cite{zhou2016fast} employed the Geman-McClure cost function and estimated the model through graduated non-convexity optimization.

Outlier rejection methods typically operate under the assumption that inliers dominate certain evaluation metrics, such as mean average error (MAE), mean square error (MSE), and inlier count. However, when the inliers rate is extremely low, as in cases of low overlap between point clouds, these methods struggle to distinguish inliers from noise, leading to registration failures. Consequently, the accuracy of point cloud registration is significantly influenced by the quality of the initial correspondences. Having better policies to generate correspondences with more inliers can enhance the outcomes of outlier rejection. Notably, within the aforementioned studies, there is almost no specific mention of the matching policy choice or comparisons between different policies in the ablation study. 

\subsection{Feature matching policy}
More recently, several deep-learned methods \cite{10076895, yu2021cofinet} estimated the transformation in an end-to-end way. These deep-learned-based methods used specific matching policies as network modules to improve the performance of registration results. Both CoFiNet \cite{yu2021cofinet}  and GeoTransformer \cite{10076895} utilized the Sinkhorn algorithm for local dense points feature matching. However, this policy is only for local matching instead of global. Similar feature matching policies are observed in other domains. In 2D-2D data association fields, the representative SuperGlue \cite{sarlin2020superglue} used Graph Neural Network to extract features and then applied the Sinkhorn algorithm to find correspondences and reject non-matched points. Notably, compared to the 2D-2D task, 3D registration needs to handle a larger number of points, especially for the Scene-level lidar scans. In non-rigid 3D registration, \cite{zhang2022searching} deployed the Hungarian algorithm \cite{kuhn1955hungarian} to ensure one-to-one matching, achieving improved matching precision. However, it only considers the object-level point clouds. The Hungarian algorithm may become extremely time-consuming when the input point clouds are large. In addition, for the low overlap scene, the one-to-one principle may also not be effective. 

In summary, these approaches treat the feature matching task as an Assignment problem and address it using relevant techniques (e.g., Sinkhorn and Hungarian algorithms). However, these methods only consider local or object-level points cloud and are not suitable for the conventional PCR method. Additionally, the transition to the Assignment problem is not reliable in general correspondence-based PCR. In the following \cref{sec:method}, we will provide a detailed explanation for this limitation and introduce a heuristic stable matching policy.

\section{Method}
\label{sec:method}
In this section, we will first present the conventional pipeline of PCR and then focus on the feature matching module. Specifically, in \cref{subsec:AP_problem}, we mainly elucidate the connection between feature matching task and the Assignment Problem as well as the Subset Assignment Problem. In \cref{subsec:GS-Matching}, we introduce the GS-Matching policy for feature matching in general PCR. In \cref{subsec:prob_analyze}, we further analyze the feature matching task using the probability theory.
\begin{figure*}[!t]
  \centering
  \includegraphics[width=0.98\linewidth]{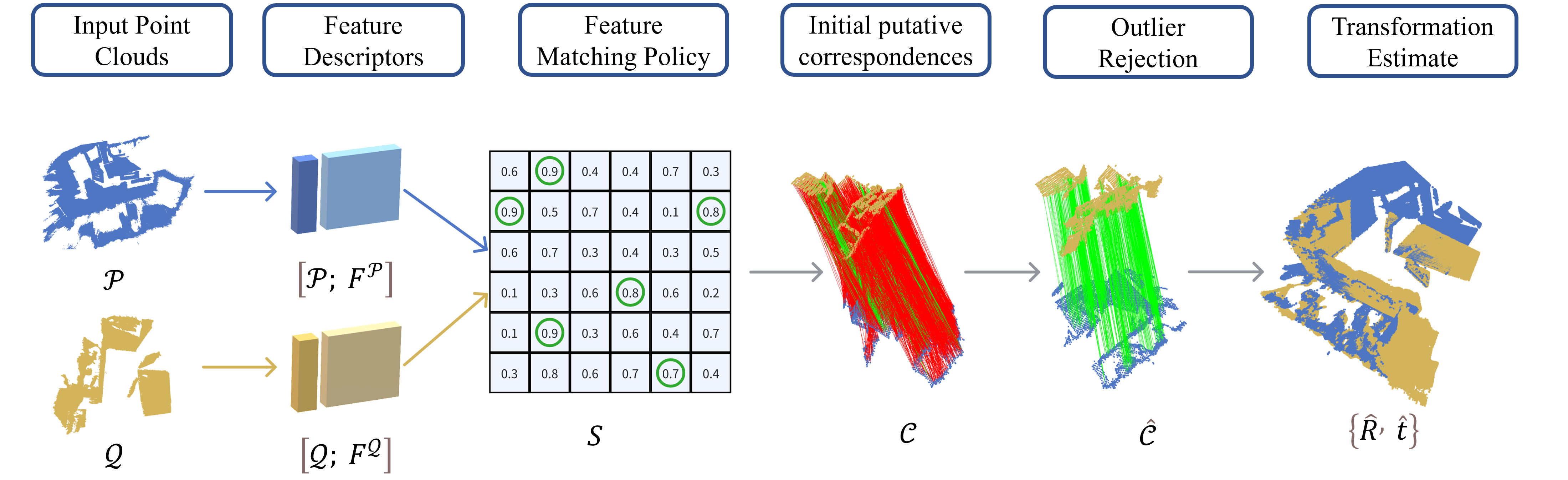}
  \caption{\textbf{The overall pipeline of the general correspondence-based PCR .} Our work focuses on the Feature Matching Policy module of this pipeline.}
  \label{fig:overall_pipeline}
\end{figure*}

\subsection{Overall pipeline}
The overall pipeline of the conventional correspondence-based PCR method is shown in figure \ref{fig:overall_pipeline}. 
Given two point clouds to be aligned: source point cloud $\mathcal{P} = \{ \bm{p_i} \in \mathbb{R}^3\ |\ i = 1, ..., m \}$ and target point cloud $\mathcal{Q} = \{ \bm{q_j} \in \mathbb{R}^3\ |\ j = 1, ..., n \}$ from a pair of partially overlapping 3D point clouds. Each keypoint has an associated local feature descriptor $\mathbf{F}_{i}^{\mathcal{P}}$ or $\mathbf{F}_{j}^{\mathcal{Q}}$ for differentiation. The initial putative correspondence set $\mathcal{C}$ can be generated via a specific matching policy in the feature space. The objective is to recover an optimal 3D rigid transformation with rotation $\mathbf{\hat{R}} \in \mathcal{SO}(3)$ and translation $\mathbf{\hat{t}} \in \mathbb{R}^3$, which minimize the following $L_2$-error $E$:
\begin{equation}
E(\mathbf{R},\mathbf{t}) = \sum_{(\bm{p}_{x_i}^{*},\bm{q}_{y_j}^{*}) \in \mathcal{C}^*} ||\mathbf{R} \cdot \bm{p}_{x_i}^{*}+ \mathbf{t} - \bm{q}_{y_j}^{*}||_{2}^{2}.
  \label{eq:L2_error}
\end{equation}
Here $\mathcal{C}^*$ stands for the ground-truth inlier set of the initial correspondences $\mathcal{C}$. Given ground truth $\mathbf{{R^*}}$ and $\mathbf{{t^*}}$, the true inlier set $\mathcal{C}^*$ can be formed as:
\begin{equation}
\mathcal{C}^* = \{(\bm{p}_{x_i},\bm{q}_{y_j})\in \mathcal{C}\ | \ (||\mathbf{R^*} \cdot \bm{p}_{x_i}+ \mathbf{t^*} - \bm{q}_{y_j}||_{2} < \tau)\}.
  \label{eq:inliers}
\end{equation}
Here $\tau$ denotes the inlier threshold. In the \cref{eq:inliers}, the true inlier set $\mathcal{C^*}$ depends on the ground truth $\mathbf{{R^*}}$ and $\mathbf{{t^*}}$ which are unknown. Since it is unable to get the true inliers in advance, the outlier rejection module becomes essential to identify a reasonable inlier set, denoted as $\mathcal{\hat{C}}$, from the initial set $\mathcal{C}$. Subsequently, we can substitute the predicted inlier set $\mathcal{\hat{C}}$ into the  $\mathcal{C^*}$ in the \cref{eq:L2_error} to compute the $\mathbf{\hat{R}}$ and $\mathbf{\hat{t}}$.
Common evaluation metrics for determining the predicted inlier set $\mathcal{\hat{C}}$ include inlier count, mean square error (MSE), etc. However, when the inlier rate is extremely low, the noise in the initial correspondences may overshadow the true inlier set under these metrics. Under this condition, none of the outlier rejection methods can effectively distinguish the true inlier set. So the initial correspondence set determines the limit of the outlier rejection module.

\subsection{Assignment Problem analyzing}
\label{subsec:AP_problem}
\subsubsection{Problem Formulation.} Given source point cloud $\mathcal{P}$ and target point cloud $\mathcal{Q}$ as well as associated feature descriptors $\{\mathbf{F}_{i}^{\mathcal{P}}\}$ and $\{\mathbf{F}_{j}^{\mathcal{Q}}\}$, the similarity matrix $\mathbf{S} \in \mathbb{R}^{m\times n}$ can be formed by computing cosine similarity between descriptors:
\begin{equation}
s_{i,j} = \frac{\langle\mathbf{F}_{i}^{\mathcal{P}},\mathbf{F}_{j}^{\mathcal{Q}}\rangle}{||\mathbf{F}_{i}^{\mathcal{P}}||_2\cdot||\mathbf{F}_{j}^{\mathcal{Q}}||_2}.
  \label{eq:cosine_similarity}
\end{equation}
Here, the $\langle\cdot,\cdot\rangle$ stands for the inner product. We can consider the $s_{i,j}$ as the cost or profit of assigning $\bm{p_i}$ to $\bm{q_j}$. Let permutation matrix $\mathbf{P}$ be a $m \times n$ binary matrix, where $p_{i,j} = 1$ if source point $x_i$ is assigned to target point $y_j$, and $p_{i,j} = 0$ otherwise. We can consider the feature matching as an Assignment Problem and the optimal permutation matrix $\mathbf{P}^*$ can be obtained by
\begin{equation}
  \mathbf{P}^* = arg max\ trace(\mathbf{P}\mathbf{S}^\text{T}), 
  \label{eq:important}
\end{equation}
where $\mathbf{P}$ is subject to:
\begin{equation}
\begin{aligned}
\mathbf{P}\in [0,1]^{m \times n}: \mathbf{P}\mathbbm{1}_n=\mathbbm{1}_m\ \text{and}\ \mathbf{P}^\text{T}\mathbbm{1}_m=\mathbbm{1}_n.
  \label{eq:important}
\end{aligned}
\end{equation}
Here, $\mathbbm{1}_n$ is the $n$ dimensional vector of ones.

\subsubsection{Solution to the Assignment Problem.} To resolve the AP, One representative exact algorithm is the Hungarian algorithm \cite{kuhn1955hungarian}, which finds the optimal solution in $O(n^3)$ time complexity. When dealing with a large number of points, the Hungarian algorithm will be very slow, rendering it unsuitable for Scene-level PCR. To address this limitation, Kantorovich Relaxation can be applied to transform the binary matrix $\mathbf{P}$ into a continuous doubly stochastic matrix $\mathbf{\Tilde{P}}$:
\begin{equation}
\begin{aligned}
\mathbf{\Tilde{P}}\in \mathbb{R}_{+}^{m \times n}: \mathbf{\Tilde{P}}\mathbbm{1}_n=\mathbbm{1}_m\ \text{and}\ \mathbf{\Tilde{P}}^\text{T}\mathbbm{1}_m=\mathbbm{1}_n,
  \label{eq:important}
\end{aligned}
\end{equation}
This relaxation reformulates the original problem as a differentiable optimal transport problem, which can be resolved efficiently by the Sinkhorn algorithm \cite{cuturi2013sinkhorn}. Once the optimal $\mathbf{\Tilde{P}}$ is obtained, the assignment can be estimated using the mutual maximum selection. 

\subsubsection{Analysis of the Assignment Problem.} Solving the Assignment Problem yields an optimal global assignment that guarantees one-to-one matching. However,  this approach assumes all the source points in $\mathcal{P}$ and target points in $\mathcal{Q}$ have a correct correspondence. This assumption holds true only when the point clouds $\mathcal{P}$ and $\mathcal{Q}$ are completely overlapped. For the PCR task, it is more common that point clouds are partially overlapped.

\subsection{GS-Matching policy}
\label{subsec:GS-Matching}
Our policy is inspired by the Gale-Shaply (GS) algorithm \cite{gale1962college}, which is well-known for finding stable matching. A stable matching refers to a situation in which no two pairs of elements would prefer to be with each other rather than their current match. In the context of Point Cloud Registration, stable matching implies that two points are not paired as correspondences simply because there exists another point more suitable for one of them. The stable matching principle can protect the matching policy from selecting an impractical matching result. Therefore, this principle is general and suits situations where the overlapping region is unknown.

\subsubsection{Analysis of the GS-Matching.} For the GS algorithm, the time complexity is $O(n^2)$ \cite{gale1962college}. However, the GS algorithm needs every row and column of the score matrix to have its own preferred list. So additional $O(n^2log(n))$ is needed to perform the sorting for every row and column. This makes the GS algorithm inefficient in the PCR task. Therefore, we improve the GS algorithm considering the time complexity and the characteristics of the PCR feature matching task. 

\begin{algorithm}[!t]
  \caption{GS-Matching Algorithm}
  \label{alg:GS_Matching}
  \KwIn{Score Matrix $\mathbf{S}$, Iteration number $K$, threshold $T_1, T_2$}
  \KwOut{Correspondences $\mathbf{corr}=\{src,tgt\}$}
  \tcp{Step 1. Generate weighted $\mathbf{S}$}
  Get the binary matrix $\mathbf{S}_{bin}$ of $\mathbf{S}$ according to $T_1$\\
  Get noise count vector $V_{NC}$ for src points by counting every row of $\mathbf{S}_{bin}$ \\
  Discard src points whose noise count is greater than $T_2$ \\
  Get priority vector $V_{weight}$ according to $V_{NC}$ \\
  Get the weighted score matrix $\mathbf{S}_{weighted}$ by applying $V_{weight}$ to each src point\\
  \tcp{Step 2. GS-Matching}
  Generate $K$ preferred candidate list for every src point and target point\\
  \While{$iter \leq K$}{
    Each src point matches with the most preferred target point\\
    Each tgt point matches with the most preferred source point\\
    save the \{src,tgt\} to $\mathbf{corr}$ only when they both prefer each other\\
    Update $\mathbf{S}_{weighted}$ by deleting the matched src and tgt points.\\
    }
  \tcp{Step 3. Nearest Neighbor step}
  \textbf{for} src points and tgt points remain:\\
  \ \ Each src point matches with the most preferred remaining tgt points\\
  \ \ save the \{src,tgt\} to $\mathbf{corr}$\\
  return $\mathbf{corr}$
\end{algorithm}

The algorithm of the source-optimal GS-Matching policy is shown as in \cref{alg:GS_Matching}.
Here the time complexity of the mutual nearest neighbor function in step 2 is $O(n^2)$ \cite{oron2017best}. Therefore our algorithm time complexity is $O(Kn^2)$. Since the $K$ we set is small and unrelated to $n$, the time complexity is also $O(n^2)$, which is much faster than the original Gale-Shapley algorithm. Compared with the GS algorithm, our matching policy has the following characteristics: 1. We perform the nearest neighbor policy for the points not matching in the GS-Matching step. If points cannot secure stable matching in this step, we infer that these points have a low possibility in the overlap region, and thus, we resort to the nearest neighbor policy for them.  2. Besides the feature score $s_{ij}$, the GS-Matching also considers the noise count as a priority for each source point. The priority increases with a decrease in the noise count. We set this priority according to the following section.

\subsection{Probability analyzing for feature matching}
\label{subsec:prob_analyze}
\begin{figure}[tb]
  \centering
  \begin{subfigure}{0.49\linewidth}
    \includegraphics[width=1.56in]{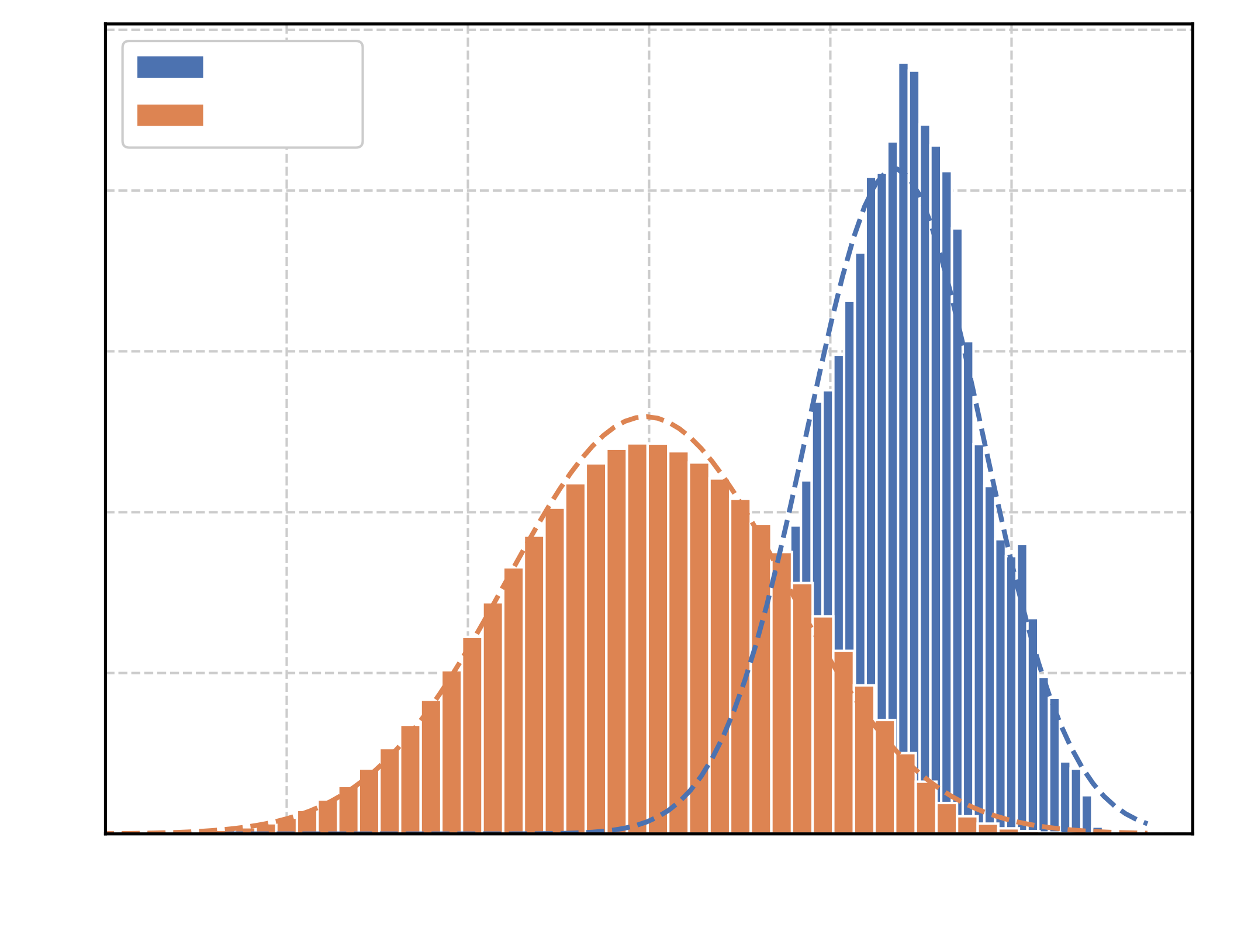}
    \caption{Statistics of Feature Similarity Score.}
    \label{fig:Probability-a}
  \end{subfigure}
  \hfill
  \begin{subfigure}{0.48\linewidth}
    \includegraphics[width=1.6in]{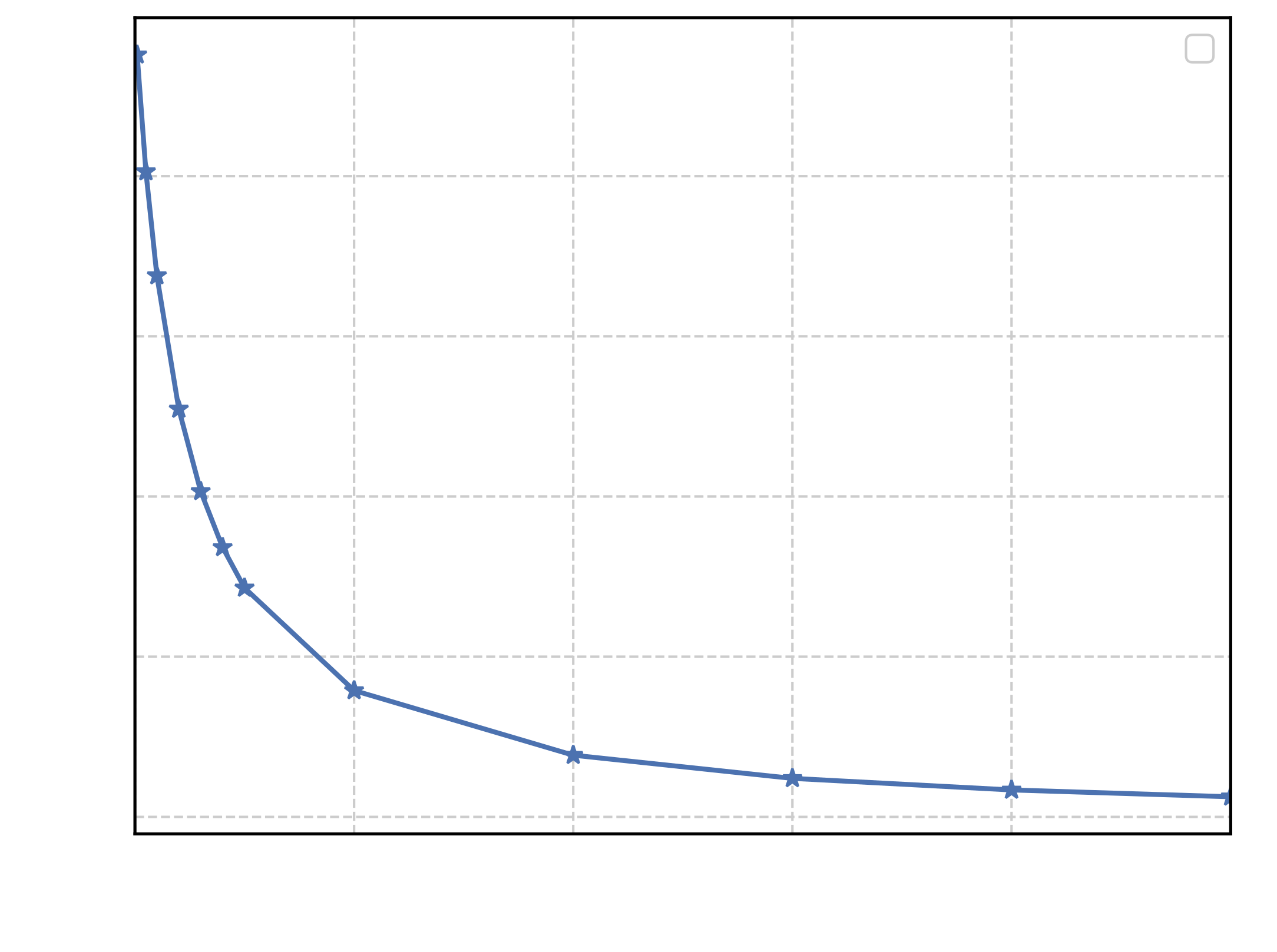}
    \caption{The probability of choosing the inlier.}
    \label{fig:Probability-b}
  \end{subfigure}
  \caption{\textbf{Probability analyzing.} For the \cref{fig:Probability-b}, we set the $m=10$ and $n=size*m$.}
  \label{fig:Probability}
\end{figure}

All feature matching policies are typically based on the assumption that the highest feature similarity score represents the highest probability of being the inlier. In this section, we further analyze how strong is this assumption. The \cref{fig:Probability-a} is the statistical data of one pair in 3dMatch using the FCGF \cite{choy2019fully} descriptor. If the distance between the transformed source point and the target point is lower than 10cm, then this pair is deemed as inlier.
We can assume the feature score of inliers $X \sim \mathcal{N}(\mu_1, \sigma^2_1)$, feature score of outliers $Y \sim \mathcal{N}(\mu_2, \sigma^2_2)$. Now we have one source point matching with all the target points, and assume there exists the correct correspondences in target points.
Then we can divide these target points into inliers set $\mathcal{X}=\{X_1, X_2,..., X_m\}$ and outliers set $\mathcal{Y}=\{Y_1, Y_2,..., Y_n\}$ and all these scores are independent and identically distributed with $X$ or $Y$.  Following the largest feature score principle, we want to get the probability the matching one is inlier:

\begin{equation}
  \text{P}(\max(\mathcal{X})> \max(\mathcal{Y})).
  \label{eq:P_initial}
\end{equation}

To calculate this probability, we first need to get the Cumulative Distribution Function (CDF) of the $X_{max} = \max(\mathcal{X})$ and $Y_{max} = \max(\mathcal{Y})$.

\begin{align}
    F_{max}(x) =\text{P}(X_{max} < x)=\prod_{i=0}^m F(x)=F^m(x).
    \label{eq:CDF}
\end{align}

Then we can get the Probability Density Function (PDF) by taking the derivative of $F_{max}(x)$:

\begin{align}
    f_{max}(x) = F_{max}^{'}(x) = mF^{m-1}(x)f(x),
    \label{eq:PDF}
\end{align}
where the $F(x)$ and $f(x)$ represents the CDF and PDF of $\mathcal{N}(\mu_1, \sigma^2_1)$. We can also compute the PDF $f_{max}(y)$ for $Y_{max}$ using the same way. Then we can get the probability in \cref{eq:P_initial}:

\begin{align}
    \text{P}(\max(\mathcal{X})> \max(\mathcal{Y})) &= \text{P}(X_{max}> Y_{max}) \\
    &= \int_{-1}^1 \int_y^1 f_{max}(x)f_{max}(y) \,dx \,dy
    \label{eq:calculate_P}
\end{align}
The \cref{eq:calculate_P} is only dependent on $m,n$, the results are shown in the \cref{fig:Probability-b}.
It is clear that the probability decreases significantly with the increase of outliers.
If we already know there exist $k>>m$ potential inliers ($score>\mu_1$), the \cref{eq:P_initial} transfer to a conditional probability:
\begin{align}
    \text{P}(\max(\mathcal{X})> \max(\mathcal{Y})\ |\ \mathcal{Y}[k]>\mu_1)
    \label{eq:conditional_p}.
\end{align}
Here, the $\mathcal{Y}[k]$ reprents the $k_{th}$ largest value of $\mathcal{Y}$. Since it is difficult to calculate this probability directly, we quantitatively analyze this issue instead. When the $m=n=1$, we can easily compute the \cref{eq:P_initial} is 0.934 and \cref{eq:conditional_p} is 0.589. For the condition $m=1,k=n$, we can infer the ratio of \cref{eq:P_initial} to \cref{eq:conditional_p} will continue to decrease along with $n$. We make a further hypothesis under the general condition $k>threshold$ when the highest feature score can not represent the inlier at all. Then we can reduce the size of points and give priority to points according to the probability analysis. 



\section{Experiment}
In this section, we mainly evaluate the impact of the GS-Matching policy on registration performance. We substitute the "Feature descriptors" module and "Outlier Rejection" module in the pipeline \cref{fig:overall_pipeline} with different descriptors and outlier rejection methods to fully and fairly test whether our matching policy is suitable for current mainstream PCR methods.

\subsection{Datasets and Experimental Setup}
\textbf{Datasets and Descriptors.} We validate our methods on three datasets, i.e., the indoor datasets 3DMatch \cite{zeng20173dmatch} \& 3DLoMatch \cite{huang2021predator} (low-overlap version of 3DMatch), and the outdoor dataset KITTI \cite{geiger2012we}. We follow the \cite{chen2022sc2} to evaluate 1623 \& 1781 partially overlapped point cloud fragments. For KITTI, we follow \cite{zhang20233d, bai2021pointdsc} and obtain 555 pairs of point clouds for testing.

\begin{table*}[!h]
  \centering
  \small
  \begin{tabular}{lccccccccccccc}
    \toprule
    & & \multicolumn{2}{c}{3DMatch FPFH} & \multicolumn{2}{c}{3DMatch FCGF} & \multicolumn{2}{c}{3DLoMatch FPFH} & \multicolumn{2}{c}{3DLoMatch FCGF} & \\
    \cmidrule(lr){3-4} \cmidrule(lr){5-6} \cmidrule(lr){7-8} \cmidrule(lr){9-10}
     & &  & RE(deg) &  & RE(deg) &  & RE(deg) &  & RE(deg) & Time\\
     & & RR(\%) & /TE(cm) & RR(\%) & /TE(cm) & RR(\%) & /TE(cm) & RR(\%) & /TE(cm) &(s)\\
    \midrule
    \multicolumn{10}{l}{\textbf{Deep Learned}} \\
    3DRegNet \cite{pais20203dregnet} & & 26.31 & 3.75/9.60 & 77.76 & 2.74/8.13 & - & -/- & - & -/- & 0.07\\
    DGR \cite{choy2020deep}& & 32.84 & 2.45/7.53 & 88.85 & 2.28/7.02 & 19.88 & 5.07/13.53 & 43.80 & 4.17/10.82 & 1.53\\
    DHVR \cite{lee2021deep}& & 67.10 & 2.78/7.84 & 91.93 & 2.25/7.08 & - & -/- & 54.41 & 4.14/12.56 & 3.92\\
    PointDSC \cite{bai2021pointdsc} & & 77.39 & \textbf{2.05/6.43} & 92.85 & 2.05/{6.50} & 27.74 & 4.11/10.45 & 55.36 & 3.79/{10.37} & 0.10\\
    \midrule
    \multicolumn{10}{l}{\textbf{Traditional}} \\
    SM \cite{leordeanu2005spectral} & & 55.88 & 2.94/8.15 & 86.57 & 2.29/7.07 & 6.06 & 6.19/12.62 & 33.52 & 4.28/11.01 & \textbf{0.03}\\
    RANSAC \cite{fischler1981random} & & 65.29 & 3.52/10.98 & 89.62 & 2.50/7.55 & 15.34 & 6.05/13.74 & 46.38 & 5.00/13.11 & 0.97\\
    GC-RANSAC \cite{barath2018graph}& & 71.97 & 2.43/7.03 & 89.53 & 2.25/6.93 & 17.46 & 4.43/10.75 & 41.83 & 3.90/10.44 & 0.55\\
    TEASER \cite{yang2020teaser}& & 75.79 & 2.43/7.24 & 87.62 & 2.38/7.44 & 25.88 & 4.83/11.71 & 42.22 & 4.65/12.07 & 0.07\\
    FGR \cite{zhou2016fast}& & 40.91 & 4.96/10.25 & 78.93 & 2.90/8.41 & - & -/- & 19.99 & 5.28/12.98 &0.89\\
    SC2 \cite{chen2022sc2} & & 83.26 & {2.09}/6.66 & 93.16 & 2.09/6.51 & 38.46 & 4.04/{10.32} & 58.62 & 3.79/{10.37} &0.11\\ 	 			 
    MAC \cite{zhang20233d}& & 83.92 & 2.11/6.79 & 93.72 & {2.03}/6.53 & 41.27 & 4.06/10.64 & 60.08 & {3.75}/10.60 &1.87\\
    SC2-Plus \cite{chen2023sc}& & 87.18 & 2.10/{6.64} & {94.15} & 2.04/{6.50} & 41.27 & \textbf{3.86/{10.06}}  & 61.15 & {3.72}/10.56 &0.28 \\
    SVC (w.o GS) \cite{zhang2024sightviewconstraintrobust} & & {88.66} & 2.18/6.87 & {94.58} & 2.07/6.60 & {45.76} & 4.04/10.62 & {67.77} & 3.93/10.83 &0.25 \\
    \midrule
    Ours & & \textbf{90.82} & 2.19/6.96 &\textbf{95.01} & \textbf{1.99/6.31} & \textbf{52.39} & {3.99}/10.35 & \textbf{71.70} & \textbf{3.71}/\textbf{10.37} & 0.20\\
    
    \bottomrule
  \end{tabular}
  \caption{Quantitative Results on 3DMatch \& 3DLoMatch dataset.}
  \label{tab:results_on_3DMatch}
\end{table*}

\textbf{Evaluation Criteria.}
Following \cite{zhang20233d, bai2021pointdsc}, we first report the
registration recall (RR) under an error threshold. For the indoor scenes, the threshold is set to (15 deg, 30 cm), while the threshold for outdoor scenes is (5 deg, 60 cm). For a pair of point clouds, we calculate the errors of translation and rotation estimation separately. We compute the isotropic
rotation error (RE) and L2 translation error (TE) as
\begin{equation}
RE(\mathbf{\hat{R}})=arccos \frac{trace(\mathbf{\hat{R}^T}\mathbf{R^*})-1}{2},\ TE(\mathbf{\hat{t}})=||\mathbf{\hat{t}}-\mathbf{t^*}||_2.
  \label{eq:error_metric}
\end{equation}
Here $\mathbf{R^*}$ and $\mathbf{t^*}$ denote the ground-truth rotation and translation.

\textbf{Implementation Details.}
Since the GS-Matching is only to generate initial correspondence, it can serve as a plug-and-play module to be combined with existing PCR methods. In the following experiments, we followed the latest PCR method \cite{zhang2024sightviewconstraintrobust} to combine the GS-Matching with SC2 \cite{chen2022sc2} and SVC \cite{zhang2024sightviewconstraintrobust}. To show the impact of the GS-Matching method on the registration results, we also show the results without GS-Matching combined. All experiments were conducted on an Intel i7-12650H CPU and NVIDIA RTX4060 GPU.

\begin{table}[!ht]
\centering
\small

\begin{tabular}{lcccc}
\toprule
& \multicolumn{2}{c}{RR(\%)} & \multicolumn{2}{c}{TE(cm)} \\
    & Ori & Ours & Ori & Ours \\
    \midrule
\textit{(i) FPFH Result} &  &  &  &  \\
SM & 6.06 & \textbf{7.86}(+1.80) & \textbf{12.62}
 & 12.73 \\
GC-RANSAC & 17.46 
 & \textbf{24.14}(+6.68) & 10.75 & \textbf{10.61} \\
TEASER & 25.88  & \textbf{31.50}(+5.62)  & 11.71& \textbf{11.03} \\
PointDSC & 27.74  & \textbf{34.14}(+6.40)  & 10.45 & \textbf{10.14} \\
SC2-PCR & 38.46 & \textbf{42.00}(+3.54)  & 10.32 & \textbf{10.17} \\
MAC & 41.27  & \textbf{44.81}(+3.54)  & 10.64 & \textbf{10.32} \\
\midrule
\textit{(i) FCGF Result} &  &  &  &  \\
SM & 33.52 & \textbf{38.63}(+5.11) & 11.01 & \textbf{10.90} \\
GC-RANSAC & 41.83 & \textbf{46.38}(+4.55) & 10.44 & \textbf{10.43} \\
TEASER & 42.22 & \textbf{48.01}(+5.79) & 12.07 & \textbf{11.11} \\
PointDSC & 55.36 & \textbf{58.90}(+3.54) & 10.37 & \textbf{10.06} \\
SC2-PCR & 58.62 & \textbf{59.74}(+1.12) & 10.37 & \textbf{9.92} \\
MAC & 60.08 & \textbf{62.27}(+2.19) & 10.60 & \textbf{10.21} \\
\bottomrule
\end{tabular}
\caption{Combined with different PCR methods}
\label{tab:comparison_methods}
\end{table}

\subsection{Results on 3DMatch \& 3DLoMatch}
We mainly combine our method with the latest PCR method SVC \cite{zhang2024sightviewconstraintrobust} to further improve its registration performance as shown in \cref{tab:results_on_3DMatch}. We also test the performance combined with multiple outlier rejection methods, including SM \cite{leordeanu2005spectral}, RANSAC \cite{fischler1981random}, GC-RANSAC \cite{barath2018graph}, TEASER++ \cite{yang2020teaser}, PointDSC \cite{bai2021pointdsc}, MAC \cite{zhang20233d}. Quantitative results are shown in \cref{tab:comparison_methods}. To be consistent with these methods in their respective original papers, we only take the Nearest neighbors policy as the original policy to compare with the GS-Matching policy.

According to the results, the following conclusions can be made: 1) The GS-Matching policy shows effectiveness for both FPFH and FCGF descriptors. For FPFH, our policy improves the best registration recall from 88.66\% \& 45.76 to 90.82\% \& 52.39\%, respectively. For FCGF is from 94.58\% \& 67.77\% to 95.01\% \& 71.70\%. 2) According to the \cref{tab:comparison_methods} The GS-Matching policy is more suitable for most outlier rejection methods. Most outlier rejection methods show improved registration performance when combined with the GS-Matching policy. 3) The GS-Matching policy can get lower rotation error (RE) and translation error (TE) in general. Even when registration recall is similar, the GS-Matching policy can get lower RE and TE metrics and we think this is mainly due to the one-to-one matching principle. 4) The GS-Matching can reduce the processing time slightly. Despite GS-Matching requiring more time for the feature-matching procedure, the whole procedure time is slightly reduced from 0.25s to 0.20s. This reduction is primarily attributed to our ability to decrease the point cloud size, as discussed in \cref{subsec:prob_analyze}. For example, for the 3DLoMatch dataset, the GS-Matching can reduce the size of initial correspondence from around 3900 to around 3200 without compromising registration performance.

\textbf{Combined with other learning descriptors.} Our method can serve as a plug-and-play module to boost the performance of existing advanced learning-based methods. In this paper, we only test our method with PREDATOR \cite{huang2021predator} and GeoTransformer \cite{qin2022geometric} on the 3DLoMatch dataset. Specifically, for the GeoTransformer, we directly use descriptors of fine points to generate initial correspondence. As shown in \cref{tab:learning_descriptors}, our method can also boost the registration recall performance of advanced learning descriptors. For the GeoTransformer method, our method achieves 82.43\% registration recall on the challenging 3DLoMatch dataset.

\begin{table}[!h]
  \centering
  
  \begin{tabular}{lccc}
    \toprule
    & \multicolumn{3}{c}{3DLoMatch Dataset} \\
    \cmidrule(lr){2-4} 
    Method & RR(\%) & RE(deg) & TE(cm)  \\
    \midrule
    PREDATOR (+PointDSC) & 68.89 & \bf{3.43} & \bf{9.60} \\
    +SC2 & 69.68 & 3.46 & 9.66\\
    +SVC (w.o GS) & 72.43 & 3.55 & 9.79\\ 
    +Ours & \bf{74.40} & {3.54} & {9.89} \\
    \midrule
    GeoTransformer (+LGR) & 75.00 & \bf{2.94} & 9.10 \\
    +SC2 & 78.11 & 3.01 & 8.69\\ 	 	 	 	 
    +SVC (w.o GS) & 81.02 & 3.09 & \bf{8.56} \\
    Ours  & \bf{82.43} & {3.01} & {8.58}\\
    \bottomrule
  \end{tabular}
  \caption{Registration results with other learning descriptors.}
  \label{tab:learning_descriptors}
\end{table}


\begin{table}[!t]
  \centering

  \small 
  \begin{tabular}{lcccc}
    \toprule
    & \multicolumn{2}{c}{RR(\%)} & \multicolumn{2}{c}{RE(º)/TE(cm)} \\
    & Ori & Ours & Ori & Ours \\
    \midrule
    \textit{i) FPFH Result}\\
    SM & 96.12 & \textbf{96.58} & 0.39\ /\ \textbf{7.76} & \textbf{0.37}\ /\ 7.94 \\
    RANSAC-100K & \textbf{90.45} & 90.09 & 1.35\ /\ 28.08 & \textbf{1.24\ /\ 26.54} \\
    TEASER++ & 91.17 & \textbf{97.66} & 1.03\ /\ 17.98 & \textbf{0.66\ /\ 11.85} \\
    PointDSC & \underline{99.64} & \underline{99.64} & 0.31\ /\ 7.08 & \textbf{0.28\ /\ 6.62} \\
    SC2-PCR & \underline{99.82} & \underline{99.82} & 0.36\ /\ 8.06 & \textbf{0.32\ /\ 7.36} \\
    \midrule
    \textit{ii) FCGF Result}\\
    SM & 95.50 & \textbf{96.22} & 0.65\ /\ 23.04 & \textbf{0.55\ /\ 20.77} \\
    RANSAC-100K & 88.29 & \textbf{98.38} & 1.28\ /\ 28.44 & \textbf{0.38\ /\ 22.28} \\
    TEASER++ & 94.96 & \textbf{95.14} & \textbf{0.38\ /\ 13.69} & 0.64\ /\ 22.89 \\
    PointDSC & \textbf{98.20} & 98.02 & 0.33\ /\ 21.11 & \textbf{0.32\ /\ 20.45} \\
    SC2-PCR & 98.20 & \textbf{98.38} & 0.33\ /\ 20.76 & \textbf{0.33\ /\ 20.32} \\
    \bottomrule
  \end{tabular}
  \caption{Registration results on KITTI dataset.}
  \label{tab:KITTI_result}
\end{table}

\subsection{Results on KITTI Dataset}
To evaluate the generability of the GS-Matching policy in outdoor environments, we conduct another experiment on the KITTI dataset \cite{geiger2012we}.\\
\textbf{Combined with different Descriptors and Outlier Rejection methods.} To be consistent with the indoor dataset experiments, we also use both FPFH and FCGF descriptors. For the outlier rejection methods, we select the representative methods of different types, including SM, RANSAC, TEASER++, PointDSC, and SC2-PCR.  

\cref{tab:KITTI_result} shows the quantitative results on the KITTI datasets. The previous method already achieved a very high registration recall performance (over 99\%), as seen our method can achieve the same best registration recall with lower RE and TE metrics. The registration experiments on the indoor scene and outdoor scene datasets consistently verify that the GS-Matching policy holds good generalization ability in different environments.


\begin{figure}[!b]
  \centering
  \includegraphics[width=2.5in]{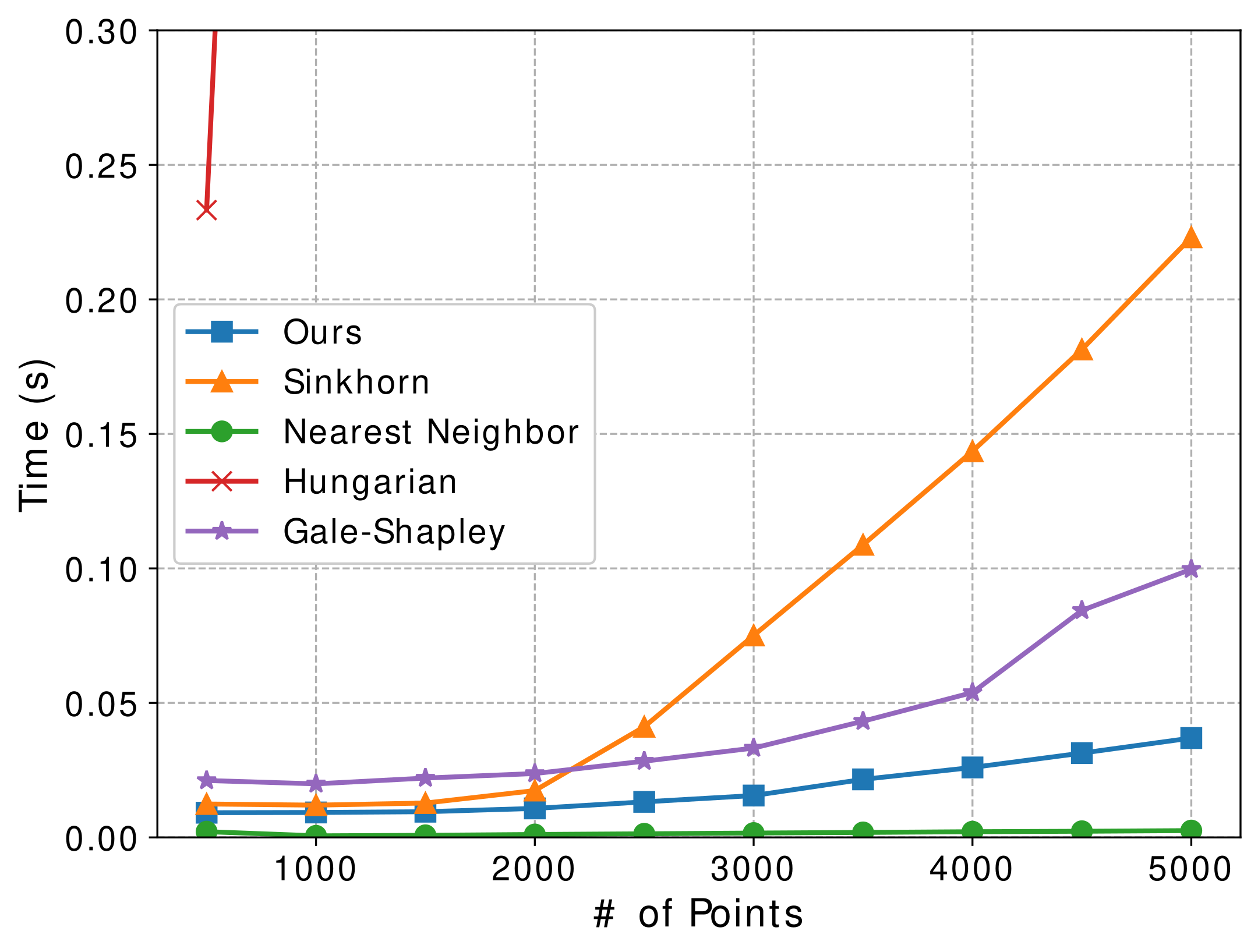}

   \caption{{Time comparison between different policies.}}
   \label{fig:time_comparision}
\end{figure}

\begin{figure*}[!t]
  \centering
  \begin{subfigure}{0.48\linewidth}
    \includegraphics[width=3in]{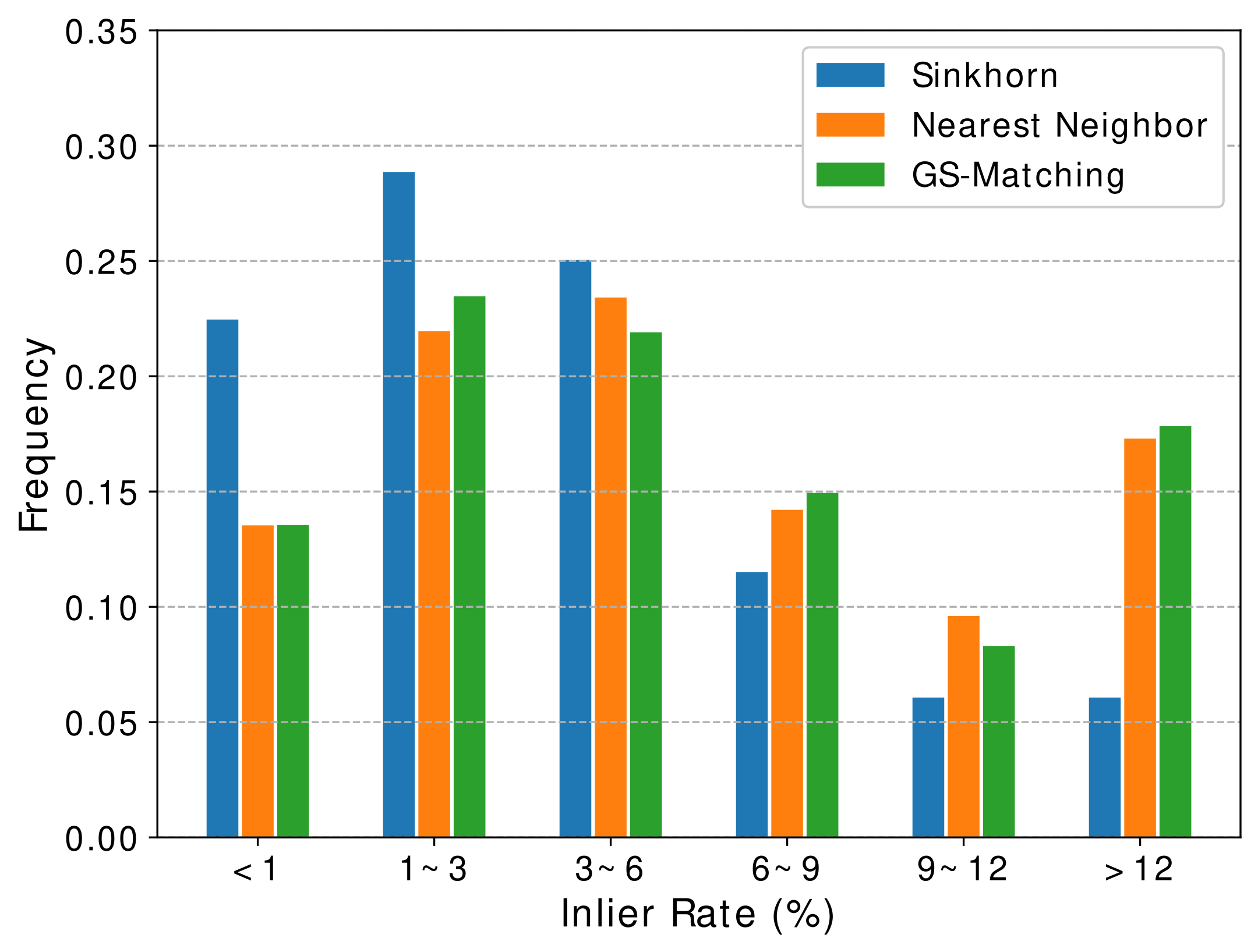}
    \caption{\textbf{The Inlier rate of different matching policies.}}
    \label{fig:Inlier_a}
  \end{subfigure}
  \hfill
  \begin{subfigure}{0.48\linewidth}
    \includegraphics[width=3in]{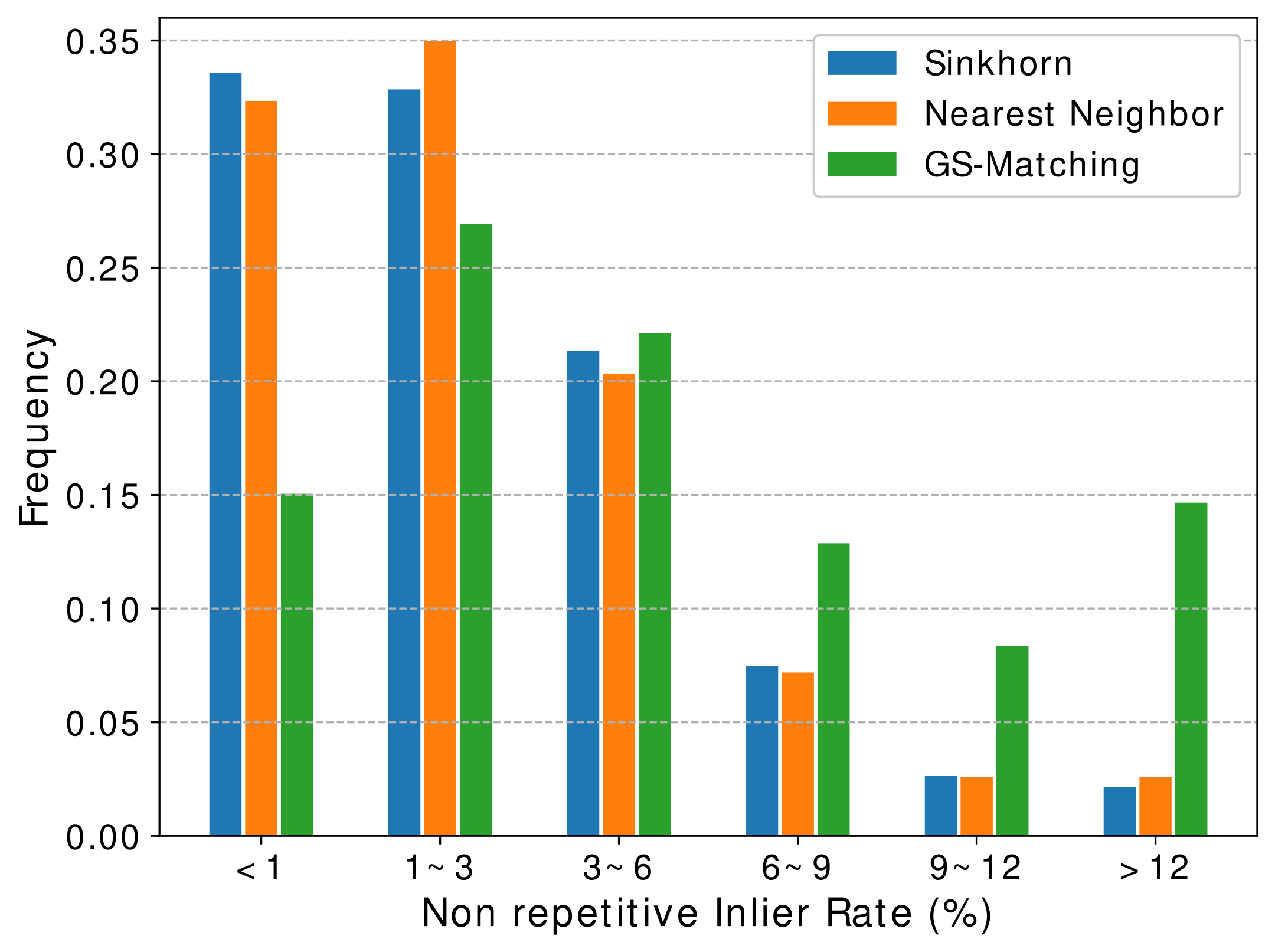}
    \caption{\textbf{The Non-repetitive Inlier rate of different matching policies.}}
    \label{fig:Inlier_b}
  \end{subfigure}
  \caption{\textbf{The comparison of Inlier Rate and Non-repetitive Inlier Rate across various matching policies.} A superior matching policy is indicated by a higher proportion of high IR or NIR.}
  \label{fig:Inlier}
\end{figure*}

\subsection{Analysis Experiments}
\label{sec:Analysis_experiments}
In this section, we conduct analysis experiments to evaluate the performance of the GS-Matching policy. For feature matching policies, we mainly focus on three evaluation metrics: 1) time complexity, 2) inlier ratio, and 3) non-repetitive inlier ratio. \\
\textbf{Time complexity analysis.} We evaluate the time complexity of various matching policies, such as the Nearest Neighbor, Sinkhorn algorithm, Hungarian algorithm, original Gale-Shapley algorithm, and the GS-Matching policy. To ensure fairness in comparison, we implement all these algorithms on the same GPU. Additionally, for the Sinkhorn algorithm, we utilize the Geomloss library \cite{feydy2019interpolating} with an entropy regularization factor set to $0.001$.

    

As shown in the \cref{fig:time_comparision}, the Hungarian algorithm exhibits slow performance, rendering it unsuitable for the PCR task. The Sinkhorn computes efficiently when the input \# of points is lower than 2000. However, the computation time increases rapidly when the \# is larger than 2500. The original Gale-Shapley algorithm performs better than Sinkhorn when the \# of points is larger than 2500. The GS-Matching further accelerates the speed based on the Gale-Shapley algorithm and the matching time is lower than 0.05s when the \# is 5000. For reference, the simplest Nearest Neighbor achieves the fastest time $\approx$0.003s with 5000 points.

\textbf{Inlier Ratio and non-repetitive Inlier Ratio.} We compare the inlier rate of the putative initial correspondences $\mathcal{C}$ using different matching policies on the 3DLoMatch Datasets. The Inlier Rate (IR) is defined as:
\begin{equation}
IR = \frac{\text{corr}\ \#\ \text{of}\ \mathcal{C}^*}{\text{corr}\ \#\ of\ \mathcal{C}}.
  \label{eq:IR}
\end{equation}
Here, the "corr" stands for correspondence, and the $\mathcal{C}^*$ stands for inliers set of correspondences and is defined by \cref{eq:inliers}. Besides IR, we also defined the Non-repetitive Inlier Ratio (NIR) as:
\begin{equation}
NIR = \frac{\text{non}\ \text{repetitive}\ \text{corr}\ \#\ \text{of}\ \mathcal{C}^*}{\text{corr}\ \#\ of\ \mathcal{C}}.
  \label{eq:NIR}
\end{equation}
Here, the non-repetitive means either $p_i$ or $q_j$ of the correspondence $(p_i,q_j)$ is not repetitive in other correspondences. The NIR is more reliable and can better indicate how many different points are involved in estimating the final transformation.

The results of IR and NIR are presented in the \cref{fig:Inlier_a} and \cref{fig:Inlier_b}. The Sinkhorn algorithm gets the worst IR and NIR results. As we discussed in the \cref{subsec:AP_problem}, the core reason is the feature matching task is not suitable to transfer to the Assignment problem under partial overlap conditions. The distributions of the IR of the Nearest Neighbor and GS-Matching are similar, but the distribution of the NIR of the Nearest Neighbor left skewed significantly due to the repeat matching problem that we discussed in \cref{sec:intro}. The NIR can better reflect the quality of the inliers, we deem that our registration results, including registration recall and RE \& TE, outperform the original method mainly because the NIR is much better. 

\section{Conclusion}
In this paper, we analyze the problem of existing feature-matching policies in the PCR field. By analyzing the feature matching task, we propose the GS-Matching policy which obeys the stable matching principle to generate better initial correspondences. Compared with other matching policies, the GS-Matching can perform efficiently and generate more reliable inliers. We also attempt to employ probabilistic methods to comprehend the feature-matching task and, based on this, eliminate some redundant points. Extensive experiments demonstrate that our method can boost the performance of mainstream PCR methods, including the state-of-the-art method, on multiple datasets.


\end{document}